\documentclass[10pt,twocolumn,letterpaper]{IEEEtran}

\usepackage[algo2e]{algorithm2e} %

\usepackage{graphicx}
\usepackage{amsmath}
\usepackage{amssymb}
\usepackage{booktabs, nicefrac}

\usepackage{color}

\usepackage{multirow,xcolor}

\usepackage{enumitem,eucal}
\usepackage{caption}
\definecolor{citecolor}{HTML}{0071bc}
\usepackage[pagebackref,breaklinks=true,letterpaper=true,colorlinks,citecolor=citecolor,bookmarks=false]{hyperref}

\usepackage[capitalize]{cleveref}
\crefname{section}{Sec.}{Secs.}
\Crefname{section}{Section}{Sections}
\Crefname{table}{Table}{Tables}
\crefname{table}{Table}{Tables}

\usepackage{threeparttable}
\usepackage{booktabs}
\usepackage{dcolumn}
\usepackage{float} 
\usepackage{subfig}
\usepackage{enumitem}
\usepackage{mathrsfs}  
\usepackage{gensymb}
\usepackage{ragged2e}

\usepackage{amsmath,amssymb,amsthm}
\usepackage{algorithm,algorithmicx,algpseudocode}
\usepackage{bm,xspace}
\usepackage{comment}
\usepackage{verbatim}
\usepackage{multirow}
\usepackage{balance}
\usepackage{url}
\usepackage{booktabs}
\usepackage{etoolbox,siunitx}
\usepackage{calc}
\usepackage{pifont,hologo}

\begin{document}

\title{Improving  Monocular Visual Odometry 
\\ 
Using Learned Depth}

 \author{
         Libo Sun$^*$,
  Wei Yin$^*$,
        Enze Xie,
        Zhengrong Li, 
        Changming Sun,
        Chunhua Shen
\thanks{
L. Sun and W. Yin claim equal contributions.
L. Sun and  W. Yin are with The University of Adelaide, Australia (e-mail: libo.sun@adelaide.edu.au; yvanwy@outlook.com). 
E. Xie is with The University of Hong Kong (e-mail: xieenze@hku.hk).
Z. Li is with Northwest A\&F University, China (e-mail: eric.lizr@outlook.com).
C. Sun is with CSIRO Data61, Australia (e-mail: Changming.Sun@data61.csiro.au).
C. Shen is with Zhejiang University, China (e-mail: chunhua@me.com).
}%
}

\markboth{Accepted to IEEE Transactions on Robotics, \today}%
{Sun \MakeLowercase{\textit{et~al.}}: 
 Monocular Visual Odometry
}

\maketitle

\begin{abstract}
Monocular visual odometry (VO) is an important task in robotics and computer vision.
Thus far,
how to build accurate and robust monocular VO systems that can work well in diverse scenarios remains largely unsolved. 
In this paper, we propose a framework to exploit monocular depth estimation for improving VO. The core of our framework is a monocular depth estimation module%
with 
a strong generalization capability for diverse scenes. It consists of two separate working modes to assist the localization and mapping. With a single monocular image input, the depth estimation module predicts a relative depth to help the localization module on improving the accuracy. With a sparse depth map and an RGB image input, the depth estimation module can generate accurate scale-consistent depth for dense mapping. Compared with current learning-based VO methods, our method demonstrates a stronger generalization ability to diverse scenes. More significantly, our framework is able to boost the performances of  existing geometry-based VO methods by a large margin.

\end{abstract}

\begin{IEEEkeywords}
SLAM, Visual Odometry, Monocular Depth Estimation
\end{IEEEkeywords}


\IEEEPARstart{M}{onocular} visual odometry (VO), which only requires monocular cameras to achieve localization and mapping, is of great importance in the field of robotics and computer vision. Monocular VO requires accurate depth to calculate camera poses and build 3D maps. However, using one camera to obtain reliable depth is a challenge problem for geometry-based VO.
As deep learning has boosted the performances of many computer vision tasks by large margins, several works~\cite{li2018undeepvo, zhan2018unsupervised,bian2019depth,sheng2019unsupervised} 
were 
proposed to introduce monocular depth estimation networks to build VO systems. 

Some methods~\cite{loo2019cnn, tateno2017cnn} propose to embed monocular depth estimation into VO systems to boost the accuracy. For example, CNN-SVO~\cite{loo2019cnn} uses depth from the monocular depth estimation model to initialize the mean and variance of the depth at a feature location. It can create reliable feature correspondences between views and achieve fast converge to the true depth. By contrast, some learning-based methods~\cite{zhan2018unsupervised, bian2019depth} propose to directly feed a video to a network to achieve both localization and mapping with a pose network to obtain the camera poses and a monocular depth estimation network to obtain the depth.

However, although impressive results can be observed from these methods~\cite{zhan2018unsupervised, loo2019cnn, bian2019depth, tateno2017cnn}, a main issue needs to be solved: 
how to make the proposed methods to be effective for diverse scenes. A common problem for current learning-based methods is that they cannot
perform zero-shot testing in both indoor and outdoor environments.  When employing a pose network instead of leveraging features and a PnP solver~\cite{lepetit2009epnp} to obtain camera poses, the robustness is of major concern. For example, we observe that pose networks used in the current learning-based frameworks cannot predict reliable poses for scenes which have complex camera motions, such as the scenes in the TUM-RGBD dataset~\cite{sturm2012evaluating}. Even though models are trained on these scenes, the pose networks still cannot predict reliable poses. Similarly, the current monocular depth estimation models lack strong generalization ability for diverse scenes and cannot provide reliable depth for VO systems.

To solve the problem mentioned above, 
we propose a framework in which we embed a robust monocular depth estimation network into a geometry-based VO system~\cite{mur2015orb} to boost the performance on localization and mapping. Compared with the current end-to-end frameworks~\cite{zhan2018unsupervised,li2018undeepvo,bian2019depth}, we do not replace any VO modules but boost the performance of VO by introducing the robustly learned depth.

Previous works~\cite{yin2020diversedepth, yin2021virtual, lasinger2019towards} have already demonstrated that without diverse high-quality RGB-D data for training,
monocular metric depth estimation
is not sufficiently robust to diverse scenes. 
Inspired by DiverseDepth~\cite{yin2021virtual}, we embed a depth estimation module that learns the affine-invariant depth in our framework. 
Our depth estimation network can work in two different modes to benefit visual odometry and mapping separately. In the first mode, the depth estimation network is fed with only an RGB image to predict a relative depth map. Although such a depth map has an unknown scale, it contains reliable near-far relationship information, which can remove outliers to maintain an accurate point set for calculating camera poses, thus boosting the localization accuracy. In the second mode, the depth network is input with an RGB image and a sparse depth map to predict a scale-consistent depth. The sparse depth from the VO stage is used for recovering the scale for all frames.  Finally, we can obtain dense and scale-consistent depths, which are applied for  dense mapping.

In summary, we propose an effective framework to combine monocular depth estimation and current VO systems. Compared with current frameworks, the proposed framework has strong generalization abilities for diverse scenes. With our proposed methods, the localization and mapping performance can be improved.
Our contributions can be summarized as follows:
\begin{itemize}
 \item We propose a monocular depth estimation method with two different working modes, which can not only generalize to diverse scenes, but also predict high-quality scale-consistent depth maps.
 
\item We
improve the accuracy of visual odometry
by 
exploiting 
the monocular depth estimation module in a 
monocular visual odometry system. 
With the help of the monocular depth estimation module, the accuracy of visual odometry can be
considerably improved, and dense 3D maps can be constructed. 
Importantly, our proposed framework can work well in diverse scenes. 
\end{itemize}

\begin{figure*}[!t]
\centering
\includegraphics[width=0.7749\textwidth]{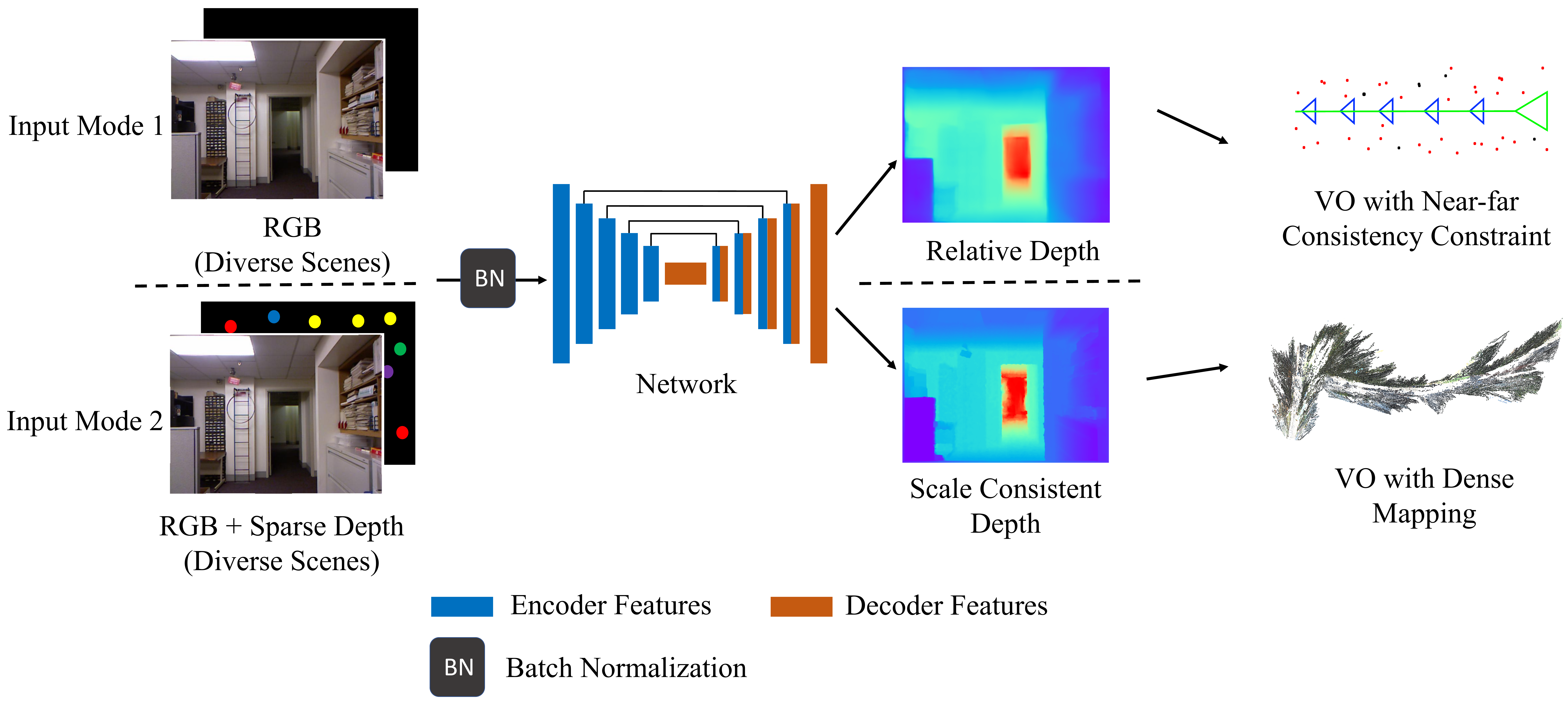}
\caption{Illustration of the framework. Our monocular depth prediction module has two input modes, \textit{i.e.}, a single RGB image or an RGB image and a sparse depth map. The relative depth map is predicted in the first mode, which can benefit the visual odometry, while the scale consistent depth map can be obtained in the second working mode. With the accurate and scale-consistent depth map, the dense 3D mapping can be performed. }
\label{fig:framework}
\end{figure*}

\section{Related Work}
\label{sec2}
\noindent\textbf{Supervised Single Image Depth Estimation.}
Monocular depth estimation is to predict the per-pixel depth from a single image, which is a long-lasting problem. Saxena et~al.~\cite{saxena2006learning} propose the first monocular depth estimation framework, which uses the Markov random field (MRF) model to explore the depth cues from multi-scale local features. Then various methods based on probabilistic models~\cite{saxena2008make3d, saxena2007learning, saxena2007depth, liu2010single} were proposed. In recent years, deep convolutional neural networks (CNN) have been used for monocular depth estimation. Eigen et~al.~\cite{eigen2014depth} propose the first multi-scale convolutional network for monocular depth prediction, in which two deep network stacks are used to obtain coarse and refined depth predictions.
Laina et~al.~\cite{laina2016deeper} propose an early state-of-the-art depth estimation network, in which they present an approach to learn feature map up-sampling and introduce the reverse Huber loss~\cite{zwald2012berhu} for optimization. More recently, to improve the geometry quality of the obtained depth, several methods~\cite{qi2018geonet, Yin2019enforcing, Wei2021CVPR} are proposed to enforce the constraints on the 3D point clouds.
These supervised depth estimation methods require ground truth depths to supervise training. The requirement of ground truth depth limits the application of supervised single image depth estimation.

\noindent\textbf{Unsupervised Single Image Depth Estimation.} Works have also been carried out on self-supervised or unsupervised learning~\cite{monodepth17, zhou2017unsupervised, bian2019depth, godard2019digging, guizilini20203d} to address the issue of the lacking of massive ground truth depth maps.
Zhou et~al.~\cite{zhou2017unsupervised} propose a framework to jointly train a single-view depth estimation network and a camera pose estimation network. The depth estimation network and the pose estimation network are trained in an end-to-end framework and can be used independently after training.
Godard et~al. propose a framework~\cite{godard2019digging} in which they use a minimum reprojection loss to handle occlusions and an auto-masking loss to ignore training pixels which violate the camera motion assumption.
Bian et~al.~\cite{bian2019depth} propose a self-discovered mask to tackle dynamic scenes to improve the depth prediction quality and the consistency between frames. 
Compared with supervised methods, these unsupervised methods do not rely on ground truth depth for training, but usually they show worse depth estimation accuracy.

\noindent\textbf{Depth Completion.}
Depth completion usually takes
a sparse depth map and an RGB image as the input to generate dense depth. Many related works~\cite{Qiu_2019_CVPR, Xu_2019_ICCV, cheng2018depth, eldesokey2018propagating, herrera2013depth, eldesokey2019confidence} have been carried out since the emergence of active depth sensors. Current depth completion methods mainly aim to complete the sparse depth maps from two types of sensors, \textit{i.e.}, structured-light scanners or LiDAR. Previous methods~\cite{kiechle2013joint, ku2018defense, mac2012patch} mainly explore hand-crafted features to fill the missing depth values. As such methods are task-specific, they usually do not show good generalization ability for diverse scenes. Recently, methods are based on deep convolutional neural networks.
Mal and Karaman~\cite{mal2018sparse} regard the sparse depth map as an additional input based depth predictor, which produces better results than the depth output from CNN with solely an image input. 
In contrast, CSPN~\cite{cheng2019learning} embeds the sampled depth in the diffusion process to make use of the sparse depth to improve the accuracy.
To improve far-distance performance and handle occlusion, Qiu et~al.~\cite{Qiu_2019_CVPR} propose a method in which attention maps and confidence masks are introduced in an end-to-end framework to obtain the completion results. 
Note that these depth completion methods take the high-quality LiDAR data as the input or reference. Such data has absolute scale and high precision, and the sparse depth points are uniformly distributed in the image.

\noindent\textbf{Geometry-based Monocular Visual Odometry.} Visual odometry and simultaneous localization and mapping (SLAM)~\cite{mur2015orb, forster2014svo, engel2017direct, engel2014lsd, newcombe2011dtam, engel2015large, kerl2013dense} are of great importance for robot navigation and mobile vision applications. According to the types of sensors used, current visual SLAM systems can be roughly divided into three categories, \textit{i.e.}, monocular, stereo, and RGB-D SLAM. Compared with stereo~\cite{engel2015large, wang2017stereo} and RGB-D SLAM~\cite{hu2012robust, kerl2013dense, whelan2015real}, monocular SLAM~\cite{klein2007parallel, mur2015orb, forster2014svo, engel2017direct} only requires one camera to achieve localization and mapping and has attracted extensive studies. 
In recent years, numerous geometry-based monocular VO frameworks have been proposed. Among these works, SVO~\cite{forster2014svo}, DSO~\cite{engel2017direct}, and ORB-SLAM~\cite{mur2015orb} are three representative geometry-based frameworks.

DSO~\cite{engel2017direct} is a typical direct VO method, which does not depend on features and can work in real-time to jointly optimize inverse depth and camera motion. DSO has good robustness performance in low-texture environments, but it usually cannot outperform feature-based methods~\cite{mur2015orb} in terms of localization accuracy in indoor environments.
By contrast, ORB-SLAM is a feature-based SLAM and needs to match feature points between images to recover depth and obtain camera poses. It obtains state-of-the-art monocular localization accuracy performance in both indoor and outdoor environments. ORB-SLAM is mainly used for localization, and it is not applicable for reconstructing dense 3D maps for related applications.
SVO is a semi-direct monocular visual odometry approach, which is a combination of feature methods and direct methods. SVO uses the sparse model-based image alignment to optimize the photometric error between frames to obtain an initial camera pose and then using bundle adjustment (BA)~\cite{triggs1999bundle} to refine the initial pose. SVO can run in real-time on onboard embedded computers and is applicable for platforms with limited computing resources, such as micro-aerial-vehicle. SVO outperforms ORB-SLAM and DSO in terms of processing speed, but does not show accuracy and robustness advantages.

\noindent\textbf{Learning-based Monocular Visual Odometry.}
With the wide application of deep learning, many learning-based monocular VO methods~\cite{wang2017deepvo, zhou2017unsupervised, li2018undeepvo, zhan2018unsupervised, sheng2019unsupervised, tiwari2020pseudo} have been proposed.
Wang et~al.~\cite{wang2017deepvo} propose an end-to-end monocular VO framework in which a deep recurrent convolutional neural network is used to predict camera poses for RGB image sequences directly. 
Liu et~al.~\cite{li2018undeepvo} propose a framework to make use of the information contained in stereo image sequences for training. Because  stereo  image  pairs  are  used  to  recover  the  scale during  training,  the  method  can  infer 6 degrees of freedom (6DoF) poses for a monocular image sequence during testing. Another framework to introduce stereo images pairs to train a monocular VO network is Depth-VO-Feat~\cite{zhan2018unsupervised}, in which spatial (left and right) and temporal (forward and backward) photometric losses are used to supervise training. In addition to the application on camera pose estimation, learning-based methods can also be used to select keyframes for VO. Sheng et~al.~\cite{sheng2019unsupervised} propose a deep network for keyframe selection and design an end-to-end unsupervised framework for monocular VO. Based on geometry-based RGB-D SLAM, Tiwari et~al.~\cite{tiwari2020pseudo} propose a pseudo RGB-D framework to couple geometry-based methods and learning-based methods. The framework uses the depth and pose generated by RGB-D SLAM to construct loss functions to train the depth estimation network, which in turn is used to generate depth to perform pseudo RGB-D SLAM. These learning-based methods have some common limitations, such as poor zero-shot testing ability, but they present approaches from different views to build monocular VO frameworks, which promotes the application of deep learning in monocular VO.

\begin{table*}[!t]
\centering
\caption{Comparison of existing RGB-D datasets. To train a robust depth estimation network, we sample training data from different types of depth datasets. These datasets are captured using different types of depth sensors in diverse scenes.}
\scalebox{1.0}{
\begin{tabular}{l|lllllll}
\toprule[1pt]
Dataset   & Diversity  &Indoor  & Outdoor & Dense & Accuracy  & Annotations & Images \\ \hline\hline
\multicolumn{7}{c}{Metric Depth Captured by RGB-D Sensors} \\ \hline \hline
NYU~\cite{silberman2012indoor}  & Low      & \checkmark &            & \checkmark & High    & RGB-D   & $407$K       \\
KITTI~\cite{geiger2013vision}   & Low      &            & \checkmark & \checkmark & High    & LiDAR   & $93$K       \\
SUN-RGBD~\cite{song2015sun}     & Low      & \checkmark &            & \checkmark & High    & RGB-D   & $10$K       \\
ScanNet~\cite{dai2017scannet}   & Low      & \checkmark &            & \checkmark & High    & RGB-D   & $2.5$M       \\
Taskonomy~\cite{zamir2018taskonomy}               & Medium   & \checkmark &            & \checkmark & High    & Laser   & $4.5$M        \\
ApolloScape~\cite{huang2018apolloscape}             & Medium   &            & \checkmark & \checkmark & High    & LiDAR   & $146$K        \\
Make3D~\cite{saxena2008make3d}  & Low      &            & \checkmark & \checkmark & High    & Laser   & $534$       \\
MegaDepth~\cite{li2018megadepth}& Medium   &            & \checkmark & \checkmark & Medium  & SFM     & $130$K       \\
DIML~\cite{cho2019large}                    & Medium   & \checkmark & \checkmark & \checkmark & Medium  & Stereo  & $2$M    \\
TUM-RGBD~\cite{sturm2012evaluating}       & Low      & \checkmark &            & \checkmark & High    & RGB-D   & $80$K   \\ 
ETH3D~\cite{schops2017multi}        & Low      &            & \checkmark & \checkmark & High    & Laser   & $454$   \\\hline\hline
\multicolumn{7}{c}{Relative Depth with Sparse Ordinal Relations} \\ \hline \hline
DIW~\cite{chen2016single}       & High     & \checkmark & \checkmark &            & Low     & Ordinal\tnote{\dag}  & $496$K  \\
Youtube3D~\cite{chen2019learning}& High    & \checkmark & \checkmark &            & Low     & Ordinal\tnote{\dag}  & $794$K \\\hline\hline
\multicolumn{7}{c}{Affine-invariant Depth} \\ \hline \hline
DiverseDepth~\cite{yin2021virtual}            & High     & \checkmark & \checkmark & \checkmark & Medium  & Stereo  & $92$K              \\
RedWeb~\cite{xian2018monocular} & High     & \checkmark & \checkmark & \checkmark & Medium  & Stereo  & $3.6$K  \\
WSVD~\cite{wang2019web}            & High     & \checkmark & \checkmark & \checkmark & Medium  & Stereo  & $1.5$M  \\
\toprule[1pt]
\end{tabular}}
\begin{tablenotes}
\footnotesize
\item[\dag]`Ordinal' means the ground truth depth map only contains ordinal relations, which are manually annotated.
\end{tablenotes}
\label{table:datasets}
\end{table*}

\section{Our
Approach}

The overall framework is illustrated in Fig.~\ref{fig:framework}. It is based on a geometry-based VO and a learning-based monocular depth estimation network. The robust depth estimation network has two working modes, which are used for promoting the localization and mapping separately. In the following subsections, we explain the whole framework in detail. 

To begin with, we introduce our depth estimation network with two working modes in Section \ref{depthModes}, Section \ref{trainingData}, and Section \ref{trainingMethod}. Then, we explain how to use the first working mode to assist localization in Section \ref{localization}. Finally, we present how to maintain scale consistency and how to achieve dense mapping by using the second mode in Section \ref{scaleRecovery} and Section \ref{secMapping}.

Before the method is described, we define the notations that are used in this paper. Throughout the paper, 
bold letters are used to denote vectors and matrices. More specifically, 
bold lowercase letters represent vectors or matrices with $1$$\times$$n$ dimensions, and bold uppercase letters represent matrices with $m$$\times$$n$ dimensions. The  uppercase Greek letters are used to denote sets. Functions, images, and frames are indicated by uppercase non-Greek letters. Scalars are represented by lowercase letters.

\subsection{Depth Network with Two Modes}
\label{depthModes}
We follow Yin et~al.~\cite{Yin2019enforcing} to build an encoder-decoder network. 
The network has two working modes, and both of them can generalize to diverse scenes. 

Firstly, when the network $\mathcal{F}$ has an input with a single RGB image, $I_{in}$, it can predict the depth map up to an affine-invariant transformation with respect to the ground truth metric depth map, \textit{i.e.}, affine-invariant depth $D_{afn}$: 
\begin{equation}
D_{afn} = \mathcal{F}(w, \mathcal{T}( I_{in} \otimes D_{o})),
\end{equation}
where $D_{o}$ is an all-zero map, $\otimes$ is the concatenation, $\mathcal{T}$ is a transformation operation, and $w$ is the network parameters. We do not enforce the model to retrieve the scale information from a single image. Instead, we disentangle the scale factor and enforce the model to learn the affine-invariant depth. Thus, the predicted depth can recover the shape of the scene. 

Furthermore, the network can predict the accurate scale consistent depth when a sparse depth map $D_{sparse}$ is available. Note that the predicted depth depends on the sparse depth and the RGB image: 
\begin{equation}
D_{metric} = \mathcal{F}(w, \mathcal{T}(I_{in} \otimes D_{sparse})).
\end{equation}
Experimentally, we find that using the batch normalization as the transformation operation $\mathcal{T}$ to unify the data distribution can achieve better performance.

\subsection{Diverse Training Data Selection}
\label{trainingData}

Many relative depth estimation methods~\cite{chen2016single, xian2018monocular} have demonstrated that diverse training data is important to improve the generalization ability of networks. Therefore, in this section, we compare the differences among current datasets and explain how we perform a selection to compose a diverse dataset for training our depth network.

We compare current popular datasets for depth estimation  in Table~\ref{table:datasets}.
There are three types of depth, \textit{i.e.}, metric depth, relative depth with sparse ordinal relations, and affine-invariant depth. Datasets with metric depth captured using depth sensors have much higher accuracy than datasets with relative depth and affine-invariant depth. However, as using depth sensors to capture such diverse datasets is very expensive, the diversity of these datasets is poor. Although several large-scale datasets are available, they only cover indoor or outdoor scenes, such as Taskonomy~\cite{zamir2018taskonomy}. In contrast, datasets with relative depth with sparse ordinal relations have the characteristics of both high diversity of scenes and large sizes, but they only contain depths with sparse ordinal relationships. Datasets with affine-invariant depth contain large-scale online stereo images with ground-truth depths obtained by using stereo matching. As the camera intrinsic parameters are unknown, the recovered depth maps are up to an affine transformation with respect to the ground truth metric depth. Although the precision is lower than that of metric depth datasets, the diversity is much higher.

As we aim to train a robust and reliable model to benefit the VO system, the selected datasets should feature both high diversity and accuracy. Therefore, we select five data sources from both datasets with metric depth and datasets with affine-invariant depth for training: Taskonomy~\cite{zamir2018taskonomy}, DIML~\cite{cho2019large}, ApolloScape~\cite{huang2018apolloscape}, DiverseDepth~\cite{yin2021virtual}, and RedWeb~\cite{xian2018monocular}. We merge DiverseDepth and RedWeb into one dataset, as they are constructed in a similar manner. In order to balance the dataset domain differences, we sample images from them to make the number of images to be around $1:1:1:1$ ratio. 

In our experiments, we found that directly merging all the data and randomly sampling images to a list of batches for training cannot obtain an appropriate model. The main reason leading to this is the large domain difference between different datasets. 
To solve this problem, we follow the curriculum learning method~\cite{hacohen2019power,yin2020diversedepth, yin2021virtual} to construct training batches. We split the whole dataset into $4$ parts, $P_{i} (i=0, ..., 3) $, based on the data collection sources, \textit{i.e.}, Taskonomy, DIML, ApolloScape, DiverseDepth, and RedWeb. We follow the multi-curriculum methods~\cite{yin2020diversedepth} to rank the training data in the order of ascending difficulty and sample the training batches in the order of ascending difficulty. In each batch, all datasets will be sampled simultaneously, and the sample proportions are the same as the dataset size proportion.

\subsection{Training Depth Estimation Network}
\label{trainingMethod}

To enrich the model with two different working modes under the same training pipeline, we propose to mix the two input modes during training. In this section, we introduce the procedure on how to sample sparse depth to train a model which can use sparse depth with different distributions and how to train the depth estimation network with two working modes.

\subsubsection{Sparse depth sampling and  normalization} Previously, several works~\cite{cheng2018depth, Qiu_2019_CVPR} have been proposed to predict dense depth from sparse depth and RGB images. These works use sparse depth from LiDAR and 
the sparse depth has a uniform distribution. However, from a VO system, we cannot obtain such a uniformly distributed sparse depth, as points in VO are usually generated based on corners and the distribution of corner points varies significantly in different images.
To make our network to use the sparse depth from VO, we use the FAST corner detector~\cite{rosten2006machine} to sample sparse depth points in our training. By using this strategy, the depth estimation network can be effective when using sparse depth with different distributions.

It is known that the sparse depth retrieved from a monocular VO system does not have a specific metric scale. The system only ensures that all the sparse depths in the keyframes share the same scale. When the working environment changes, the scale may vary. To make the sparse depth to have a similar scale in training and testing,
we normalize each sparse depth map with the largest depth of the sparse points. After concatenating it with the RGB image, they are processed by
batch normalization. 

The sampling and normalization processes are expressed as:
\begin{equation}
    D_{sparse} = \mathcal{N}(\mathcal{S}(D_{gt}, I_{in})),
\label{eq: sampling sparse depth}
\end{equation}
where $\mathcal{N}( \cdot )$ is the normalization function and  $\mathcal{S}( \cdot )$ is the sampling function in which a corner detector is used for the sampling.

\subsubsection{Training the first mode} 
The first mode aims to predict a relative depth from a single RGB image. Such a relative depth is trained as an affine-invariant depth ($z$), which requires a scale and a shift transformation to recover the metric depth $z^{*}$, \textit{i.e.}, $z^{*} = sz + t$, where $s$ is the scale and $t$ is the shift. During training, we employ the scale-shift-invariant loss proposed in~\cite{lasinger2019towards} to supervise the training for the  network. The scale-shift-invariant loss is given as follows:
\begin{equation}
\begin{aligned}
L_{ssi}=\frac{1}{2n} \sum_{i=1}^{n} \left(\textbf{z}^T_{i}\textbf{h} - z_{i}^{gt}\right)^{2} \quad {\rm with} \quad \quad \quad 
\\
\textbf{h}= \left ( \sum_{i=1}^{n}{\textbf{z}_{i}}{\textbf{z}_{i}^{T}} \right )^{-1} \left ( \sum_{i=1}^{n}{\textbf{z}_{i}}{z}_{i}^{gt} \right )
\quad {\rm and} \quad
{\textbf{z}_{i}}=\left (z_{i}, 1 \right )^{T},
\end{aligned}
\label{eq:ssil}
\end{equation}
where $n$ is the number of points, and $z^{gt}$ is the ground truth depth.

Apart from the scale-shift-invariant loss, we enforce a virtual normal loss~\cite{Yin2019enforcing} to boost the performance. To begin with, the predicted depth and the ground truth depth are unprojected into the 3D point cloud space. Then point groups are randomly and globally sampled to construct virtual planes to obtain normal vectors. Finally, the virtual normal loss which minimizes the normal difference between the predictions and the ground truth is calculated as follows:

\begin{equation}
    \def\bn{ {n} }
\label{eq:vnl}
    L_{vn} = \frac{1}{n}
    \sum_{i=1}^{n}
    \| { \textbf{v}_{i}^{pred}} -{\textbf{v}_{i}^{gt}}\|_{1},
\end{equation}
where $\|.\|_{1}$ represents the L1 norm and $\textbf{v}_{i}$ is the virtual normal recovered from the depth.

The overall loss $L_{1}$ for training the first mode is defined as:
\begin{equation} 
    \def\bn{ { n  } }
\label{eq:overall_loss}
    L_{1} = L_{ssi} + L_{vn}.
\end{equation}

\subsubsection{Training the second mode} When the model has an input with a sparse depth map, it is provided with a strong prior. Therefore, we enforce a mean square error loss ($L_{mse}$) on these sparse points:
\begin{equation}
\label{eq:mse}
    L_{mse} = \frac{1}{2n} 
    \sum_{i=1}^{n}
     ({ z_{i}^{pred}} -{z_{i}^{gt}})^2.
\end{equation}

The overall loss for training the second mode is:
\begin{equation}
    \def\bn{ { n  } }
\label{eq:overall_loss_sdepth}
    L_{2} =  L_{ssi} + L_{vn}  + L_{mse}.
\end{equation}

During training, the two input modes are selected randomly, and the loss functions are selected based on the input mode.

\begin{figure}[t]
\begin{center}
\includegraphics[width=0.499\textwidth]{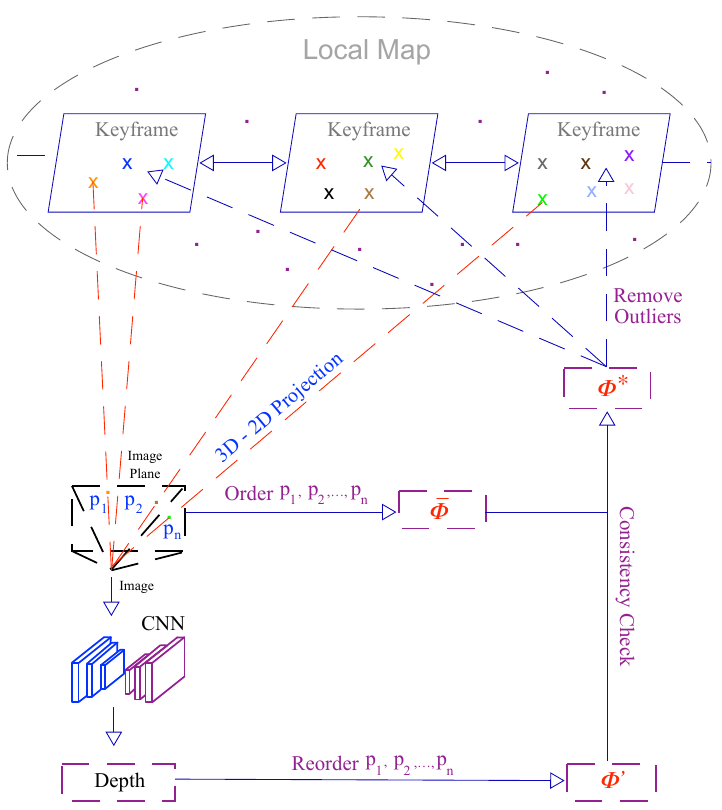}
\end{center}
\caption{Near-far consistency constraint. Points in the local map are projected onto an image plane to output $\overline{\Phi}$ which contains the near-far relations of these points. The monocular depth estimation network takes the RGB image and generates a dense depth map, which can construct another near-far relation set ${\Phi}'$. The points with different near-far weights in $\overline{\Phi}$  and ${\Phi}'$ compose the outlier set ${\Phi}^*$.}
\label{fig:consistency}
\end{figure}

\subsection{Monocular VO with Near-far Consistency Constraint}
\label{localization}
We follow the method used in ORB-SLAM~\cite{mur2015orb} to perform localization. When a new frame $F_c$ arrives, its associated camera pose is obtained by minimizing the re-projection error function:
\begin{equation}
E = \sum_{i, j}\mathcal{H}(\textbf{e}_{i, j}^T\bm{\Lambda}_{i, j}^{-1}\textbf{e}_{i, j}), 
\label{reproject1}
\end{equation}
where $\textbf{e}_{i, j}$ is the error term in which a point $i$ in $F_c$ is observed in keyframe $F_j$, $\mathcal{H}$ is the Huber robust cost function~\cite{huber1973robust}, and $\bm{\Lambda}$ is a covariance matrix. For more details about Eq.~(\ref{reproject1}), including the derivation, we refer to the work of~\cite{mur2015orb}. To improve the accuracy of visual odometry, we propose a near-far consistency constraint to remove map points which have wrong depth values. The near-far consistency check process is shown in Fig.~\ref{fig:consistency}.

For all the points in the local map, we project all of them onto the current image plane. The projected position of each 3D point $\textbf{p} = (x, y, z)$ is calculated as: 
\begin{equation}
\textbf{p}_c = \textbf{K}(\textbf{R}_{cw}\textbf{p} + \textbf{t}_{cw}),
\label{project3d} 
\end{equation}
where \textbf{K} contains the camera intrinsic parameters for the projection, $\textbf{R}_{cw}$ and $\textbf{t}_{cw}$ are the rotation and translation matrices, which transform a 3D position from the initial world coordinate system to the current camera coordinate system, and $\textbf{p}_c$ represents the projected position on the current image plane. After projecting all the local map points, we have points set $\Phi$,
which contains all the projected points on the current image plane. 

For each $\textbf{p}_h = (u_h, v_h, z_{p_h})^T$ $\in$ $\Phi$, $z_{p_h}$ is its depth value in the VO component and $(u_h, v_h)$ is its 2D position on the image plane. 
Based on the depth information in VO, we organize all the points in $\Phi$ in an ascending depth order and construct an ordered set $\overline{\Phi}$:

\begin{equation}
\begin{aligned}
\overline{\Phi} = \{ \bar{\textbf{p}}_1, \bar{\textbf{p}}_2, \bar{\textbf{p}}_3, ..., \bar{\textbf{p}}_n \} \\ 
{\rm with}  \quad
z_{\bar{p}_1} < z_{\bar{p}_2} < z_{\bar{p}_3} < \cdots < z_{\bar{p}_n}.
\label{arrange} 
\end{aligned}
\end{equation}

By using the first depth estimation mode, our depth estimation network can predict a relative depth map $D_{c}$ for the current RGB image $I_c$. For each $\textbf{p}_h = (u_h, v_h, z_{p_h})^T \in \Phi$, it has a predicted depth value $z_{p_h}'$ contained in $D_{c}$:
\begin{equation}
z_{p_h}' = D_c(u_h, v_h).
\label{depthelem}
\end{equation}
Therefore, for all the points in $\Phi$, we can reorder them to obtain a new ordered set ${\Phi}'$ by using their depth in  $D_{c}$:%

\begin{equation}
\begin{aligned}
{\Phi}' = \{\textbf{p}_{1}', \textbf{p}_{2}', \textbf{p}_{3}', ..., \textbf{p}_{n}'\}\\  {\rm with}\;  \quad
z_{p_1'}' < z_{p_2'}' < z_{p_3'}' < \cdots < z_{p_n'}'.
\end{aligned}
\label{depthArr1} 
\end{equation}

For each $\bar{\textbf{p}}_i\in\overline{\Phi}$, the index $i \in \{1, 2, 3, ..., n\}$ for this point represents the near-far relations between this point and all the other points in set $\overline{\Phi}$. Correspondingly, in set ${\Phi'}$, this point is reordered at index $j$ and represented as ${\textbf{p}_j'}$. The indices $i$ and $j$ can describe their near-far relations with other points in the ordered set $\overline{\Phi}$ and, ${\Phi'}$ respectively. Therefore, they are used to maintain the near-far consistency. 
The near-far consistency factor $\lambda_{\overline{p}_i}$, which describes the consistency between the depth from VO and the predicted depth from the depth estimation network, is defined as:
\begin{equation}
\begin{aligned}
\lambda_{\overline{p}_i} = |i - j|.
\end{aligned}
\label{nearfarc} 
\end{equation}
We define $\Phi^{*}$, which contains all the outliers that do not maintain the near-far consistency between $\overline{\Phi}$ and ${\Phi'}$. Thus, $\Phi^{*}$ and the outlier element $\textbf{p}^*_i$ in $\Phi^{*}$ are given as:
\begin{equation}
\begin{aligned}
\Phi^{*} = \{\textbf{p}^*_1, \textbf{p}^*_2, \textbf{p}^*_3, ..., \textbf{p}^*_n\}\\  {\rm with} \quad \textbf{p}^*_i\in\overline{\Phi} \quad  {\rm and} \quad   \lambda_{{p}^{*}_i} > \sigma,
\end{aligned}
\label{nearfarc_1} 
\end{equation}
where $\sigma$ is the near-far consistency threshold. All points in $\Phi^{*}$ will be removed from the VO system.

\subsection{Scale Recovery of Learned Depth}
\label{scaleRecovery}
We project sparse points from VO onto the image plane of each keyframe $F_{\tau}$ to recover the scale of the learned depth. For a point $\textbf{p}_{\tau_i} = (x_{\tau_i}, y_{\tau_i}, z_{\tau_i})^T$ observed by $F_{{\tau}}$, the projection function is defined as:

\begin{equation}
(u_{\tau_i}, v_{\tau_i}, 1)^{T} = \textbf{K}\textbf{p}'_{\tau_i} \quad {\rm with }  \quad \textbf{p}'_{\tau_i} = \frac{1}{z_{\tau_i}}\textbf{p}_{\tau_i},
\label{project1}
\end{equation}
where $(u{_{\tau_i}}, v{_{\tau_i}})$ is the projected position on the image plane of $F_{{\tau_i}}$ and $\textbf{K}$ contains the camera intrinsic  parameters. After projecting all the sparse points, we have a set $\Omega = \{\hat{\textbf{p}}_1, \hat{\textbf{p}}_2, \hat{\textbf{p}}_3, ..., \hat{\textbf{p}}_n\}$, which contains all the projected points from VO. For each $\hat{\textbf{p}}_i = (u_i, v_i, z_i)^{T} \in \Omega$,  we retrieve its learned depth value, which is denoted as $z_i{'}$ at  position $(u_i, v_i)$ of the learned depth map, and express it as $\widetilde{\textbf{p}}_{i} = (u_i, v_i, z_i{'})^{T}$. 
As each $\hat{\textbf{p}}_i \in \Omega$ can be expressed as $\widetilde{\textbf{p}}_i$, we have $\widetilde{\Omega} = \{\widetilde{\textbf{p}}_1, \widetilde{\textbf{p}}_2, \widetilde{\textbf{p}}_3, ..., \widetilde{\textbf{p}}_n\}$. 
The scale recovery is regarded as an operation to find a scale factor $\theta_{s}$ to make  $z_{i} \approx \theta_{s} * z'_i$ for each $\hat{\textbf{p}}_i \in \Omega$ and $\widetilde{\textbf{p}}_i \in \widetilde\Omega$. 

Because we hope to simplify our implementation, we use the median scale to approximate the overall scale $\theta_{s}$:
\begin{equation}
\theta_{s}\approx \mathcal{M}(z_1/z_1', z_2/z_2', z_3/z_3', ..., z_n/z_n'), \label{approximate}
\end{equation}
where $\mathcal{M}$ represents the operation to obtain the median value.
\begin{table*}[t]
\centering
\setlength{\abovecaptionskip}{10pt}
\caption{The comparison with state-of-the-art  methods on five zero-shot datasets. The Abs-Rel (\%) metric is used for the comparison. Depth in meters is used for calculation. 
`Ours-RGB' denotes that the network only takes RGB images (first mode), while `Ours-RGB+sdepth' represents taking RGB images and sparse depth (second mode). The scale of `Ours-RGB' is recovered by using ground truth depth before the evaluation. The best results on each dataset are in bold.
`$\rule{0.3cm}{0.1mm}$' represents that the method is trained on the corresponding dataset. 
}
 \setlength{\tabcolsep}{4.2pt}
\scalebox{1}{
\begin{tabular}{l|l|l|llll}
\toprule[1pt]
\multirow{3}{*}{Method}& \multirow{3}{*}{\begin{tabular}{@{}c@{}} Training\\dataset \end{tabular}} & Backbone & \multicolumn{4}{c}{Testing on zero-shot datasets} \\ \cline{3-7} 
        &  &  &  NYU & KITTI & ETH3D & ScanNet \\
        &  &  &  \multicolumn{4}{c}{Abs-Rel (\%)} \\ \hline \hline
\multicolumn{7}{c}{Learning Metric Depth + Single-scene Dataset} \\ \hline \hline
Yin et~al.~\cite{Yin2019enforcing} & NYU  & ResNeXt-101 &$\underline{10.8}$
&$35.1$  &$29.6$  &$13.7$ \\
Alhashim et~al.~\cite{alhashim2018high} & NYU & DenseNet-169 &$\underline{12.3}$   &$33.4$  &$34.5$   &$12.5$  \\
Yin et~al.~\cite{Yin2019enforcing} & KITTI & ResNeXt-101  &$26.7$  &$\underline{7.2}$  &$31.8$   &$23.5$   \\
Alhashim et~al.~\cite{alhashim2018high}& KITTI & DenseNet-169  &$23.5$  &$\underline{9.3}$   &$32.1$  &$20.5$  \\ \hline \hline
\multicolumn{6}{c}{Learning Relative Depth + Diverse-scene Dataset} \\ \hline \hline
Li and Snavely~\cite{li2018megadepth}&  MegaDepth &ResNet-50  &$19.1$  &$19.3$  &$29.0$  &$18.3$ \\

Chen et~al.~\cite{chen2016single}& DIW  & ResNeXt-50  &$16.7$ &$25.6$  &$25.7$  &$16.0$ \\
Xian et~al.~\cite{xian2018monocular}&  RedWeb   & ResNeXt-50  &$26.6$   &$44.4$    &$39.0$  &$18.2$ \\  \hline \hline
\multicolumn{6}{c}{Learning Affine-invariant Depth + Diverse-scene Dataset} \\ \hline \hline
Lasinger et~al.~\cite{lasinger2019towards} &MV + MegaDepth + RedWeb & ResNet-50 &$19.1$  &$29.0$   &${23.3}$ &$15.8$  \\
Ours-RGB & $4$ datasets  & ResNeXt-50 &${13.5}$  &${16.5}$ &$24.8$ &${11.6}$ \\ \hline \hline
\multicolumn{6}{c}{Learning Affine-invariant Depth + Diverse-scene Dataset}\\ \hline \hline
Ours-RGB+sdepth & $4$ datasets  & ResNeXt-50 &$\textbf{3.31}$  &$\textbf{5.84}$ &$\textbf{22.5}$ &$\textbf{2.92}$ \\ \toprule[1pt]
\end{tabular}}
\label{table: zero-shot comparison}
\end{table*}

\subsection{Dense Mapping Using Learned Depth}
\label{secMapping}
Previous geometry-based dense mapping methods~\cite{pizzoli2014remode} usually use a depth filter to obtain depth maps for mapping. However, such methods require small variations between frames, making them not to work on datasets with a low camera frame rate, such as the KITTI dataset. In contrast, learning-based depth estimation does not depend on depth filters and does not have the requirement for camera frame rate. 
As our framework is embedded with a generalized monoclar depth estimation network, we take the depth estimation network to produce dense depth for 3D mapping. 

To improve the mapping quality, pixel-wise intensity and depth consistency check is performed between each new keyframe $F_2$ and its nearest keyframe $F_1$. For each point $\textbf{p}$ in the initial world coordinate systems, its corresponding positions in the $F_2$  and $F_1$ camera coordinate systems are denoted as $\textbf{p}_2$ and $\textbf{p}_1$ respectively. If we define $\boldsymbol{\zeta}_2$ and $\boldsymbol{\zeta}_1$ to represent the camera poses of $F_2$  and $F_1$, the $\textbf{p}_2$ and $\textbf{p}_1$ can be given as:

\begin{equation}
\textbf{p}_1 = \textbf{R}_{1w}\textbf{p} + \textbf{t}_{1w} \quad {\rm and} \quad \textbf{p}_2 = \textbf{R}_{2w}\textbf{p} + \textbf{t}_{2w}\label{tranform1},
\end{equation}
where $\textbf{R}_{1w}$, $\textbf{R}_{2w}$, $\textbf{t}_{1w}$, and $\textbf{t}_{2w}$ are the rotation and translation parts of $\boldsymbol \zeta_{1}$ and $\boldsymbol \zeta_{2}$. For each $\textbf{p}_{2}$,  its corresponding position $\textbf{p}_{12}$ in the camera coordinate system of $F_1$ can be obtained by rewriting Eq. (\ref{tranform1}) as:

\begin{equation}
\textbf{p}_{12} = \textbf{R}_{1w}\textbf{R}^{-1}_{2w}\textbf{p}_2 + \textbf{R}_{1w}\textbf{t}_{w2} + \textbf{t}_{1w}\label{tranform2}
\end{equation}
\begin{equation}
\textbf{t}_{w2} = -\textbf{R}^{-1}_{2w}\textbf{t}_{2w} \label{tranform}.
\end{equation}

For each pixel whose position on the image plane of $F_2$ is  $(u_2, v_2)$, using our depth estimation network can predict its depth $z_2$. Therefore, its 3D position $\textbf{p}_2$ in the camera coordinate system of $F_2$ is:
 \begin{equation}
\textbf{p}_2 = \textbf{K}^{-1}\textbf{p}_{2}' \quad {\rm with} \quad \textbf{p}_{2}' = (u_2z_2, v_2z_2, z_2)^{{T}},   \label{tranform3}
\end{equation}
where $\textbf{K}^{-1}$ is the inverse of camera intrinsic matrix $\textbf{K}$. For each pixel in the image of $F_2$, we can obtain its 3D position $\textbf{p}_{12}$ in the camera coordinate system of $F_1$ by combining Eq.~(\ref{tranform2}), Eq.~(\ref{tranform}), and Eq.~(\ref{tranform3}):
\begin{equation}
\begin{aligned}
\textbf{p}_{12} =
\textbf{R}_{1w}\textbf{R}^{-1}_{2w}\textbf{K}^{-1}(u_2z_2, v_2z_2, z_2)^{{T}} + \textbf{R}_{1w}\textbf{t}_{w2} + \textbf{t}_{1w} \\  {\rm with}  \quad
\textbf{t}_{w2} = -\textbf{R}^{-1}_{2w}\textbf{t}_{2w}. 
 \quad  \quad \quad \quad \quad  \quad
\end{aligned}
\label{depthArr} 
\end{equation}

For each transformed $\textbf{p}_{12} = (x_{12}, y_{12}, z_{12})^{T}$, its  position $(u_{12}, v_{12})$ on the image plane of $F_1$ is obtained by using the projection function:
\begin{equation}
u_{12} = \frac{x_{12}}{z_{12}}f_x  + c_x \quad {\rm and} \quad v_{12} = \frac{y_{12}}{z_{12}}f_y + c_y, \label{tranform6}
\end{equation}
where $f_x$, $f_y$, $c_x$, and $c_y$ are the camera focal lengths and principle point location. Finally, the pixel-wise depth and intensity consistency check between $F_2$ and $F_1$ are performed as follows:

\begin{equation}
| z_1 - z_{12} | < \delta  \quad {\rm and }  \quad | I_{1}(u_{12}, v_{12}) - I_2(u_2, v_2) | < \gamma,
\label{tranform7}
\end{equation}
where $z_1 = D_{1}(u_{12}, v_{12})$ is the depth value contained in the predicted depth map $D_{1}$ of $F_1$, $ I_{1}$ and $ I_{2}$ represent the images of $F_1$ and $F_2$ respectively, and $\delta$ and $\gamma$ are two thresholds to remove outliers. Points which pass the consistency check will be used for dense mapping.

\section{Experiments}
In this section, we conduct several experiments to demonstrate the effectiveness of our method. Firstly, we perform zero-shot testing on diverse datasets to evaluate the generalization ability of our depth estimation network. Then, we conduct experiments to illustrate the effectiveness of our framework to combine monocular depth estimation and VO. Finally, we show the potential of our framework to improve the performances of other VO frameworks.

\begin{figure*}[!bth]
\centering
\includegraphics[width=1\textwidth]{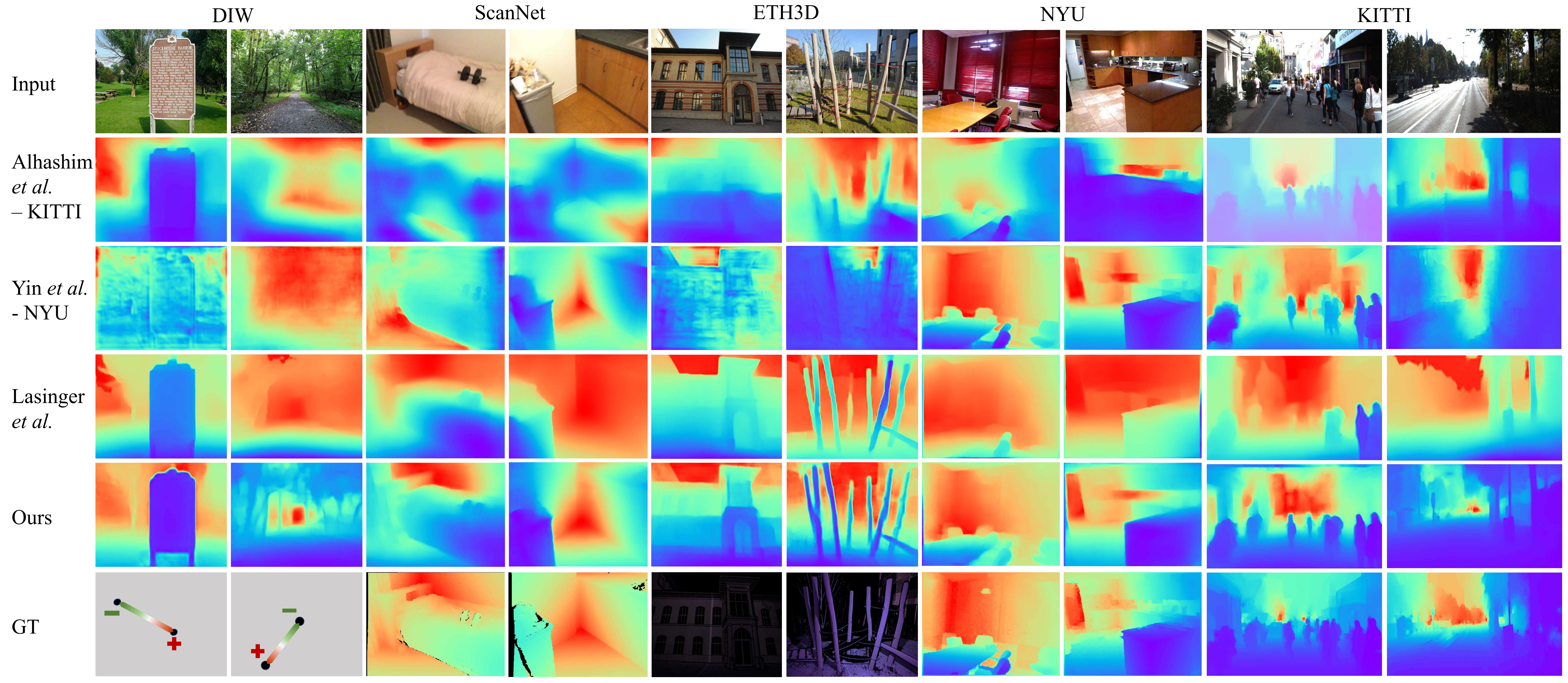}
\caption{Qualitative comparisons with previous methods on five zero-shot datasets. Learning metric depth methods (Alhashim et~al.~\cite{alhashim2018high} and Yin et~al.~\cite{Yin2019enforcing}) cannot generalize to diverse scenes on the NYU and KITTI datasets, while learning affine-invariant depth (Lasinger et~al.~\cite{lasinger2019towards}) can generalize to diverse scenes, but the details are not satisfactory. By contrast, our method (first mode) not only predicts more accurate depths on five unseen datasets, but also recovers better details on indoor and outdoor scenes.}
\label{fig:sto_cmp}
\end{figure*}

\subsection{Experiments Setups}
\noindent\textbf{Depth network implementation details.} The depth estimation network is implemented and trained by using PyTorch~\cite{paszke2019pytorch}. It is converted to a PyTorch C++ model to be embedded into our VO system after training.
We use ResNeXt-50~\cite{xie2017aggregated} as the backbone, and the weights of our network are initialized with ImageNet~\cite{deng2009imagenet} pretrained weights. A polynomial decaying method with a base learning rate $0.0005$ and a power of $0.9$ is applied for SGD optimization. The weight decay and the momentum are set to $0.0005$ and $0.9$ respectively. The batch size is 16 in our experiments.

\noindent\textbf{Depth evaluation datasets.} We perform zero-shot testing on several depth estimation datasets to evaluate the performance of our depth estimation network.
Our depth estimation network is trained on Taskonomy~\cite{zamir2018taskonomy}, DIML~\cite{cho2019large}, ApolloScape~\cite{huang2018apolloscape}, DiverseDepth~\cite{yin2021virtual}, and RedWeb~\cite{xian2018monocular}, while it is tested on NYUD-V2~\cite{silberman2012indoor}, KITTI~\cite{eigen2014depth}, ScanNet~\cite{dai2017scannet}, and ETH3D~\cite{schops2017multi}. These testing datasets contain both indoor and outdoor diverse scenes. 

\noindent\textbf{VO evaluation datasets.} To demonstrate that our proposed monocular depth prediction method can benefit current VO methods in various scenes, we conduct VO experiments on the KITTI~\cite{geiger2012we}, TUM-RGBD~\cite{sturm2012evaluating}, and RobotCar~\cite{maddern20171} datasets. Note that these testing datasets are unseen by our depth estimation network during training. 

\noindent\textbf{Evaluation metrics.} We follow previous depth estimation methods~\cite{laina2016deeper, monodepth17} to evaluate the performance of monocular depth estimation. The evaluation metrics of depth estimation are as follows:
\begin{itemize} 
\item mean absolute relative error (Abs-Rel): $\frac{1}{n}\sum_{i=1}^{n} \frac{\left | z_{i}-z_{i}^{\ast} \right |}{z_{i}^{\ast}}$;

\item square relative error  (Sq-Rel):  $\frac{1}{n}\sum_{i=1}^{n} \frac{( z_{i}-z_{i}^{\ast})^2}{z_{i}^{\ast}}$;

\item root mean squared error (RMS): $ \sqrt{\frac{1}{n}\sum_{i=1}^{n}{\left (z_{i}-z_{i}^{\ast} \right)^{2}}} $;
\item root mean squared log error (RMS-log): $\sqrt{\frac{1}{n}\sum_{i=1}^{n}{\left (\log_{10}z_{i} - \log_{10}z_{i}^{\ast} \right)^{2}}} $;
\item the accuracy under threshold ($\delta_{i} < 1.25^{i}, i=1, 2, 3$): $ \%$ of $z_{i}$  s.t. \rm{max}$\left (\frac{z_{i}}{z_{i}^{\ast}}, \frac{z_{i}^{\ast}}{z_{i}} \right ) < 1.25^{i}$.
\end{itemize}
where $z_i$ is the ground truth depth and $z_{i}^{\ast}$ represents the predicted depth.
Furthermore,  we apply absolute trajectory error (ATE)~\cite{sturm2012evaluating} to evaluate the localization accuracy of visual odometry.

\subsection{The Generalization Ability of Depth Estimation Network}

\begin{table}[t]
\caption{The illustration of the robustness of `RGB and sparse depth' input mode. The Abs-Rel (\%) metric is used for the comparison. Depth in meters is used for calculation. We illustrate the performance of our method with different sparse depth maps. `-' represents that there is no available models for the evaluation.
Our method can obtain satisfactory performances with different sampling ratios and distributions of sparse depths.}
 \setlength{\tabcolsep}{4.2pt}
\scalebox{0.9}{
\begin{tabular}{c|l|l|llll}
\toprule[1pt]
\multirow{2}{*}{\begin{tabular}[c]{@{}c@{}}Sample \\ Method\end{tabular}}    & \multicolumn{1}{c|}{\multirow{2}{*}{Testing Datasets}} & \multirow{2}{*}{Methods} & \multicolumn{4}{c}{Sample Ratio} \\
                & \multicolumn{1}{c|}{}     &     & 0\%   & 0.1\%   & 1\%   & 1.5\%   \\ \hline
\multirow{6}{*}{\begin{tabular}[c]{@{}c@{}}Uniform\\ Sample\end{tabular}}    & \multirow{3}{*}{NYU}   & CSPN~\cite{cheng2018depth}    & -  &$11.79$  & $1.41$ & $1.16$     \\
   &   & DeepLiDAR~\cite{Qiu_2019_CVPR}    &-  &-  &- &- \\
   &   & Ours    & $13.5$    & $4.03$   &$3.31$  &$3.56$    \\ \cline{2-7} 
               & \multirow{3}{*}{KITTI}      & CSPN~\cite{cheng2018depth} & -  &- &-  & -   \\
               &   & DeepLiDAR~\cite{Qiu_2019_CVPR}    &-  &$20.22$  &$4.72$ & $2.03$ \\
     &           & Ours     & 16.5      &11.4         & 6.16      & 5.84        \\ \hline
\multirow{6}{*}{\begin{tabular}[c]{@{}c@{}}Key-points\\ Sample\end{tabular}} & \multirow{3}{*}{NYU}                                  &  CSPN~\cite{cheng2018depth} &-  &$21.33$  & $5.71$  & $4.9$ \\
&   & DeepLiDAR~\cite{Qiu_2019_CVPR}    &-  &-  &- &- \\
                     &    & Ours  & $13.5$  & $5.57$ & $4.06$ & $4.01$   \\ \cline{2-7} 
   & \multirow{3}{*}{KITTI}     &  CSPN~\cite{cheng2018depth} & -  &-  &-  &-         \\
   &   & DeepLiDAR~\cite{Qiu_2019_CVPR}    & -  & $21.36$ &$7.48$ &$5.2$ \\
   &   & Ours    & $16.5$     & $13.21$  & $7.17$ & $6.32$        \\ \toprule[1pt]
\end{tabular}}
\label{table: sdepth comparison}
\end{table}

To demonstrate the generalization ability of our depth estimation network, we use $4$ unseen datasets for zero-shot testing, \textit{i.e.}, NYUD-V2, KITTI, ScanNet, and ETH3D. We compared our method with learning metric depth, learning relative depth, and learning affine-invariant depth methods on diverse scenes. The results are given in Table~\ref{table: zero-shot comparison}. 
We notice that learning metric depth can only work on a specific scene and is unable to generalize to diverse scenes. Learning relative depth methods show good generalization ability, but their Abs-Rel errors are very high on all testing datasets. By contrast, learning affine-invariant depth is better on generalization and recovering geometric information. As can be seen from Table~\ref{table: zero-shot comparison}, our method can obtain much better performance on both indoor and outdoor scenes than previous methods.

We also carry out qualitative comparisons on five unseen datasets. Some example results are shown in Fig.~\ref{fig:sto_cmp}. It is clear that our method can output much better depth maps. No matter whether it is in indoor or outdoor environments, the shape of our predicted depth is more clear. 

\subsection{Benefits of Using Sparse Depth}
Our depth estimation has two working modes. Both of these two modes have the generalization ability to work in diverse scenes. The main difference between the two modes is that the second mode can take sparse depths to recover scale and improve accuracy. This happens when accurate sparse depth is available. The sparse depth can be generated from LiDAR or VO systems.

\begin{figure}[t]
\centering
\includegraphics[width=0.489\textwidth]{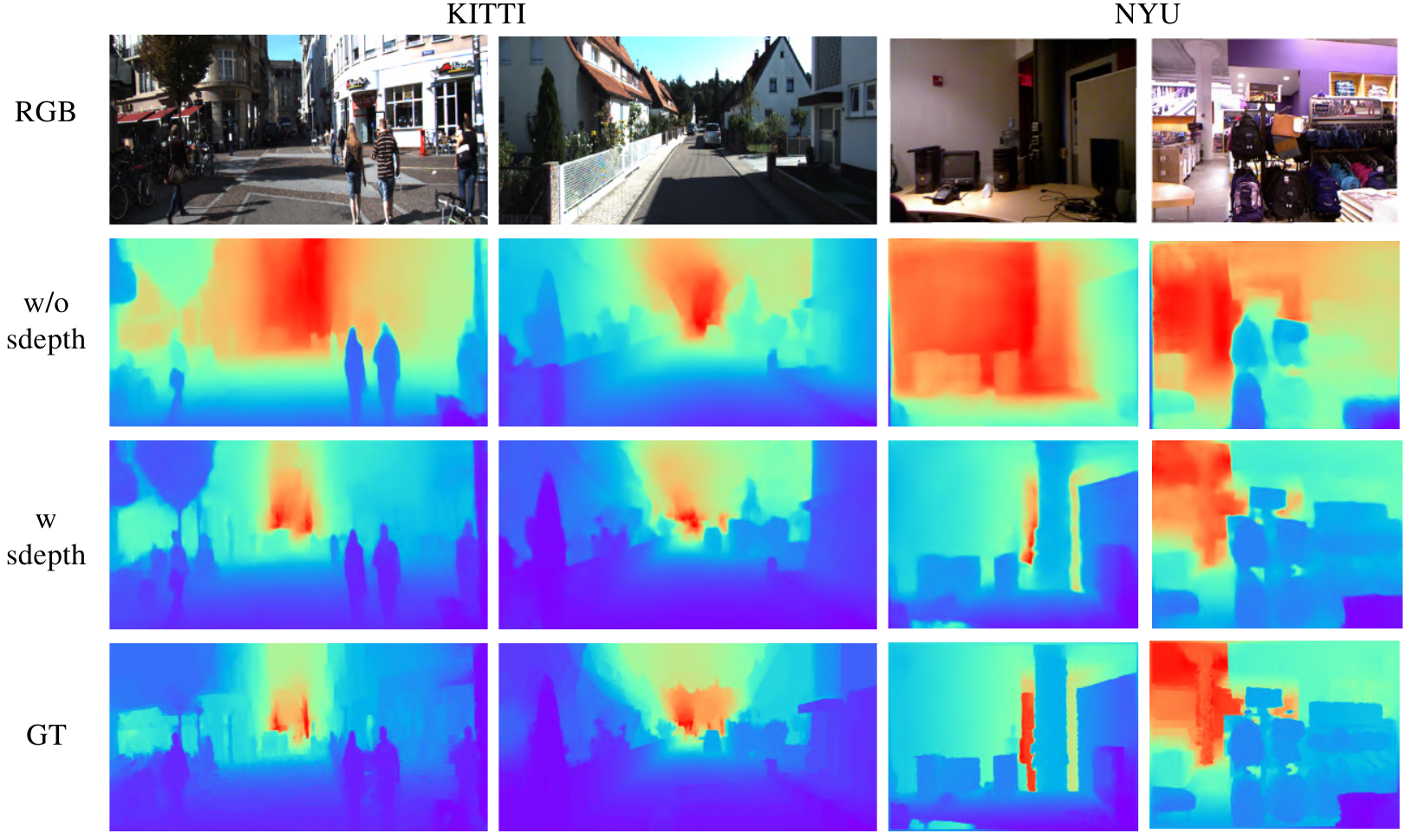}
\caption{Examples of two input modes, \textit{i.e.}, RGB images only and RGB images with sparse depth. `w/o sdepth' denotes inputting the model with only an RGB image. `w sdepth' means inputting an RGB image and a sparse depth map. It is clear that with the spars depth,  the network can significantly improve the performance. 
}
\label{fig:w_wo_cmp}
\end{figure}

Our framework is proposed for monocular VO, but it has a strong and extensive ability to work together with other depth sensors which can provide sparse depth.
We perform a wide range of experiments on KITTI and NYU to evaluate the performances when using sparse depth from LiDAR or other types of depth sensors. We compare approaches with different sampling ratios of sparse depth and different sampling methods, \textit{i.e.}, sampling sparse depth uniformly or sampling key-points.  
The results are shown in Table~\ref{table: sdepth comparison}. We can see that on both the NYUD-V2 and KITTI datasets, the performance is improved significantly if sparse depth is available. Besides, when the ratio of valid points increases in the sparse depth map, the accuracy will also increase. Furthermore, our model is robust to the distribution of sparse points. We notice that if key-points are used to sample spare depth the performance of CSPN drops significantly (from 11.79 to 21.33 on the NYU dataset), while our method can obtain comparable performance when sampling sparse points uniformly or sampling key-points. We also perform qualitative comparisons (see Fig.~\ref{fig:w_wo_cmp}) to demonstrate the importance of sparse points to recover high-quality depth. We can see that with the help of sparse depth,  the network can predict much better depth maps, especially in far-distance areas.

If sparse depth sensors are not available, we can use the depth from monocular VO as the sparse depth input. The comparison between using sparse depth from VO and without using sparse depth are shown in Table~\ref{Table:SparseCmp}. The evaluation is performed for the keyframes in the VO component only by using the setting of~\cite{monodepth17}. The sparse depth from the VO stage is very sparse (usually less than 0.1\%),
but  using it still improves the $\delta$\textless 1.25 metric from 0.782 to 0.824. 

\subsection{Visual Odometry Evaluation}
We perform VO experiments in both indoor and outdoor environments to evaluate the precision and robustness of our method. The experiments are conducted on the KITTI~\cite{geiger2012we} (outdoor), RobotCar~\cite{maddern20171} (outdoor), and TUM-RGBD~\cite{sturm2012evaluating} (indoor) datasets. For all the experiments on the KITTI and TUM-RGBD datasets, we use the setting used in ORB-SLAM~\cite{mur2015orb}. The setting on RobotCar remains the same with those on KITTI, except the camera intrinsics. We use the tool provided by~\cite{maddern20171} to generate images on RobotCar and images are resized to 1280$\times$960 for image generation.   

\begin{table}[t]
\footnotesize
\renewcommand\arraystretch{1.0}
\begin{center}
\caption{The comparison between using VO sparse depth and without using sparse depth. `RGB' represents using RGB images only. `RGB + S' indicates using RGB images and sparse depth from VO. The evaluations are performed for VO keyframes on KITTI (depth in meters). 
For Abs-Rel, Sq-Rel, RMS, and RMS-log, the lower the value, the higher the accuracy. For the $\delta$\textless 1.25 ($i = 1$) metric, the higher the value, the higher the accuracy.
}
{
\begin{tabular}{l|c|c|c|c|c} 
\toprule[1pt]
Input & Abs-Rel (\%) & Sq-Rel & RMS & RMS-log & $\delta$\textless 1.25 \\
\hline
RGB & 14.713  &   1.087  &   6.109  &   0.231  &   0.782   \\
\hline
RGB + S & 13.351  &   0.860  &   5.094  &   0.210  &   0.824 \\
\toprule[1pt]
\end{tabular}
} 
\label{Table:SparseCmp}
\end{center}
\end{table}

Note that our monocular depth estimation network has not been trained on these datasets. The absolute trajectory error (ATE) is used to evaluate the localization accuracy. Each sequence is tested for $5$ times and the mean ATE is reported. To recover the scale for evaluation, trajectories are aligned with seven degrees of freedom (7DoF) by following the alignment method used in~\cite{loo2019cnn,mur2015orb}.

The results on the KITTI, RobotCar, and TUM-RGBD datasets are shown in Table~\ref{KITcmp}, Table~\ref{robotcarCmp}, and Table~\ref{TUMCmp} respectively. The results of using ORB-SLAM and DSO are from~\cite{loo2019cnn}, and the methods that are unable to complete a sequence normally are noted as `$\times$'. Our method is based on ORB-SLAM. In these three tables, compared with the state-of-the-art geometry-based methods, \textit{i.e.}, ORB-SLAM and DSO, our method has noticeably better performance. Compared with the baseline, ORB-SLAM, our method can improve its performance in different testing scenes.

\begin{table*}[t]
\centering
\caption{Absolute keyframe trajectories  RMSE (in meters) on the KITTI dataset. Trajectories are aligned with 7DoF,
using 
the ground truth. `$\times$' denotes that the tracking is lost at some points or a significant portion of the sequence is not processed by the system. $\uparrow$ denotes our improvements to the baseline (ORB-SLAM).}
\scalebox{1}{
\begin{tabular}{l|ccccccccccc}
\toprule[1pt]
\multicolumn{1}{l|}{\multirow{2}{*}{Method}} & \multicolumn{11}{c}{Testing on KITTI (RMSE)}                                                        \\
\multicolumn{1}{r|}{}                        & $00$       & $01$ & $02$       & $03$     & $04$     & $05$      & $06$      & $07$      & $08$       & $09$      & $10$      \\ \hline
SVO~\cite{forster2014svo}                                          & $\times$          & $\times$    & $\times$         & $\times$        & $58.4$ & $\times$         & $\times$        & $\times$        & $\times$          & $\times$         & $\times$         \\
DSO~\cite{engel2017direct}                                          & $113.18$ & $\times$   & $116.81$ & $1.39$ & $0.42$ & $47.46$ & $55.61$ & $16.71$ & $111.08$ & $52.22$ & $11.09$ \\
CNN-SVO~\cite{tateno2017cnn}                                      & $17.52$  & $\times$   & $50.51$  & $3.45$ & $2.44$ & $8.15$  & $11.50$ & $6.51$  & $10.97$  & $10.68$ & $4.83$  \\
ORB(w/o Loop)~\cite{mur2015orb}                                & $77.95$  & $\times$    & $41.00$   & $1.01$ & $0.93$ & $40.35$ & $52.22$ & $16.54$ & $51.62$  & $58.17$ & $18.47$ \\

SfM-Learner~\cite{zhou2017unsupervised} & $104.87$ & $109.61$ & $185.43$ & $8.42$ & $3.10$ &$60.89$ & $52.19$ & $20.12$ & $30.97$ & $26.93$ & $24.09$\\

Depth-VO-Feat~\cite{zhan2018unsupervised} & $64.45$ & $203.44$ & $85.13$ & $21.34$ & $3.12$ & $22.15$ & $14.31$ & $15.35$ & $29.53$ & $52.12$ & $24.70$ \\ 

SC-SfMLearner~\cite{bian2019depth} & $93.04$ & $85.90$ & $70.37$ & $10.21$ & $2.97$ & $40.56$ & $12.56$ & $21.01$ & $56.15$ & $15.02$ & $20.19$\\ 
\hline

Ours                                         & $61.79^\uparrow$  & $\times$   & $28.79^\uparrow$  & $0.98^\uparrow$ & $0.89^\uparrow$ & $31.08^\uparrow$ & $45.67^\uparrow$ & $15.99^\uparrow$ & $42.09^\uparrow$  & $45.95^\uparrow$ & $8.11^\uparrow$  \\ \toprule[1pt]
\end{tabular}}
\label{KITcmp}
\end{table*}
\begin{figure*}[bth!]
\centering   
\vspace{-0.4cm}
\subfloat[] 
{
	\includegraphics[width=0.46\textwidth]{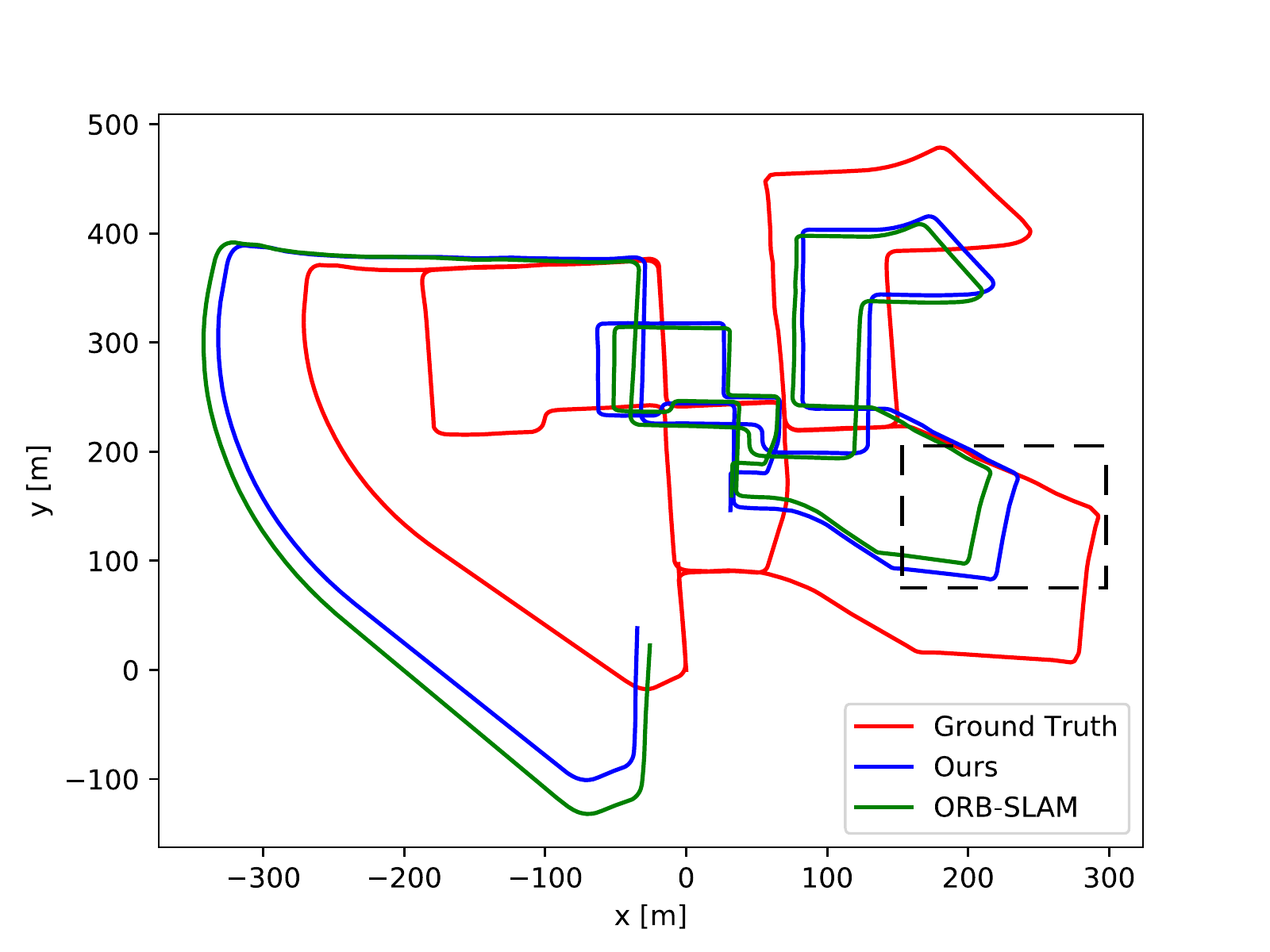}
	\label{fig:traj_s00}
}
\subfloat[] 
{   
	\includegraphics[width=0.46\textwidth]{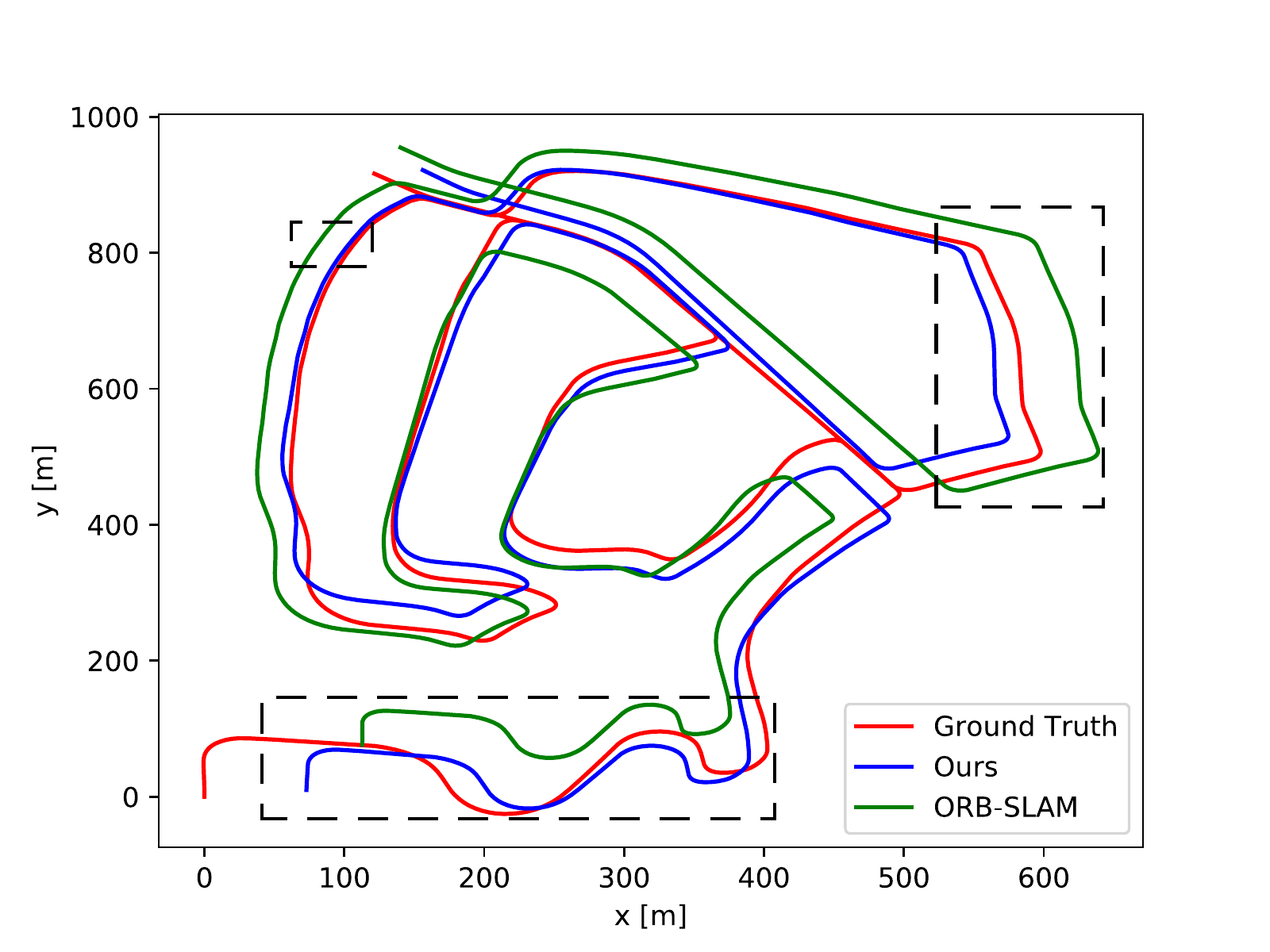} 
	\label{fig:traj_s02}
} \\
\vspace{-0.4cm}
\subfloat[] 
{
	\includegraphics[width=0.46\textwidth]{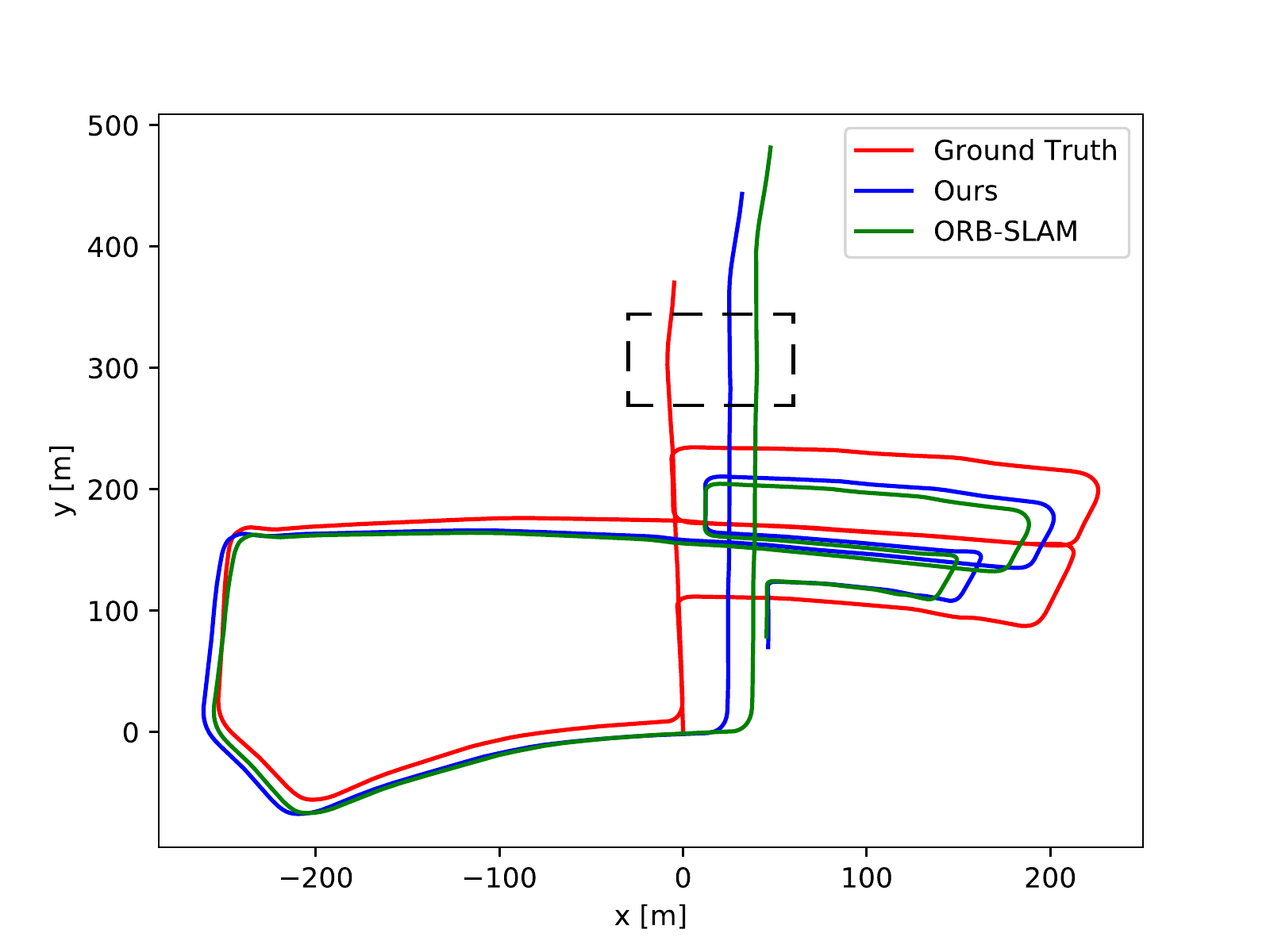}
	\label{fig:traj_s0}
}
\subfloat[] 
{ 
	\includegraphics[width=0.46\textwidth]{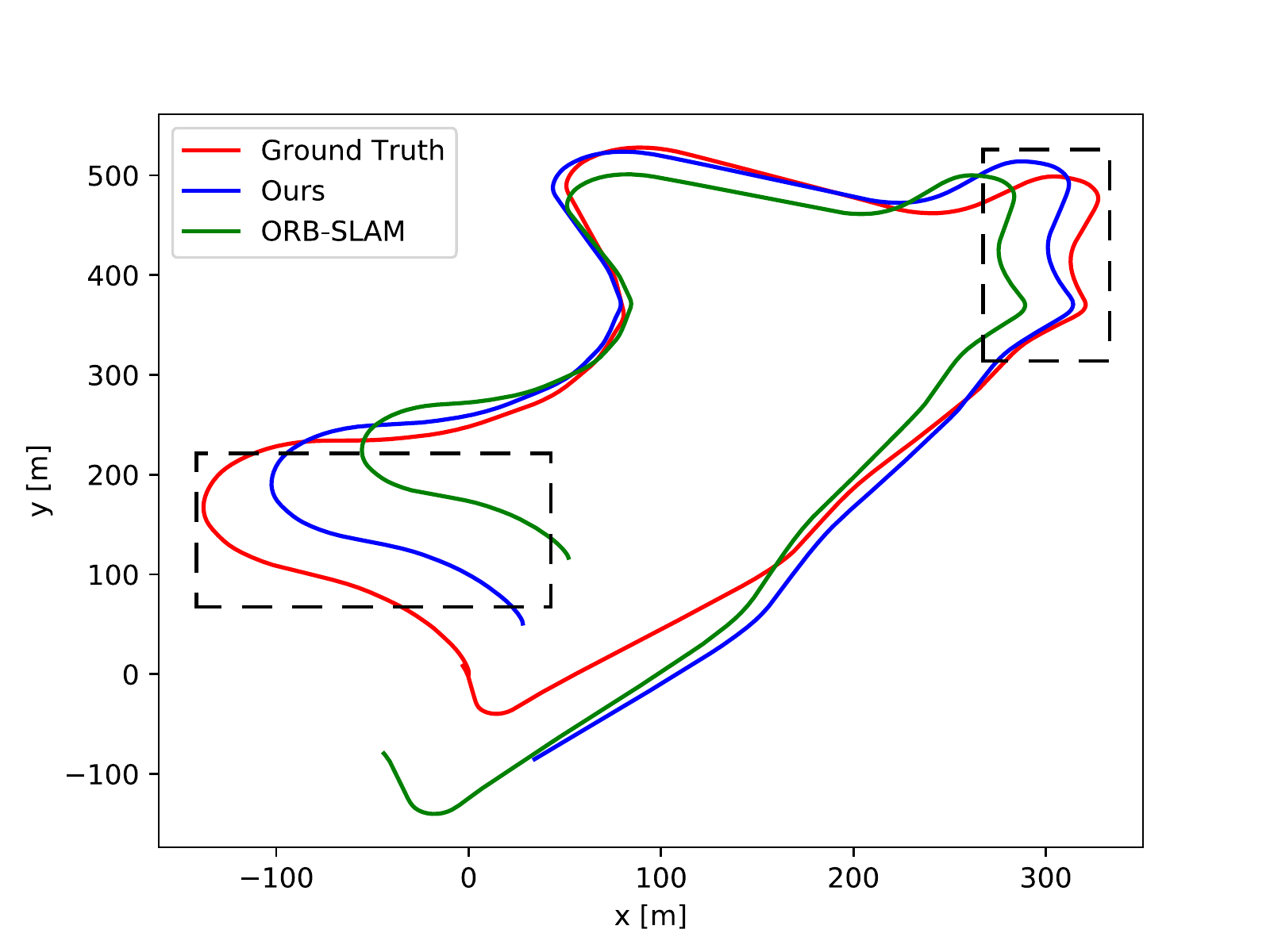} 
	\label{fig:traj_s09}
}
\caption{The comparison of trajectories between ORB-SLAM (w/o loop closure) and our method on 4 KITTI sequences. (a) sequence 00, (b) sequence 02,  (c) sequence 05,  (d) sequence 09. Some areas for trajectory comparisons are marked with black boxes. }
\label{fig:traj02} 
\end{figure*}

We can observe that on the KITTI dataset our method outperforms current state-of-the-art learning-based methods in terms of overall accuracy. As we mentioned earlier, using a pose estimation network to train a VO framework to obtain camera poses has a limitation that the current pose estimation networks are not reliable for a long sequence with diverse camera motions. Because current learning-based frameworks are mainly effective on the KITTI dataset, we do not make comparisons with them on the RobotCar and TUM-RGBD datasets.

CNN-SVO is targeted for improving the performance of SVO. It shows its accuracy advantage on the KITTI dataset, but it fails on the TUM-RGBD dataset. The main reason is that it depends on monodepth~\cite{monodepth17} which does not have a strong generalization ability for zero-shot indoor testing (mainly for outdoor city scenes).

Our framework
targets 
a different aim,
compared with that of 
CNN-SVO. We require our framework to have the generalization ability to work in diverse environments and have the potential to be applied
to several 
VO frameworks (Section \ref{DSO_Exp}). 

Besides 
the ATE comparison, qualitative comparisons between our trajectories and ground truth are illustrated in Fig.~\ref{fig:traj02}. The initial keyframe trajectories are aligned with ground truth using similarity transformation. The aligned trajectories of using ORB-SLAM and our method are placed in the same coordinate system for visual comparison. From these visual comparison results in Fig.~\ref{fig:traj02}, we can see that compared with the results of ORB-SLAM,  trajectories generated by our method are closer to the ground truth.

\begin{figure*}[]
\begin{center}
\includegraphics[width=1\textwidth]{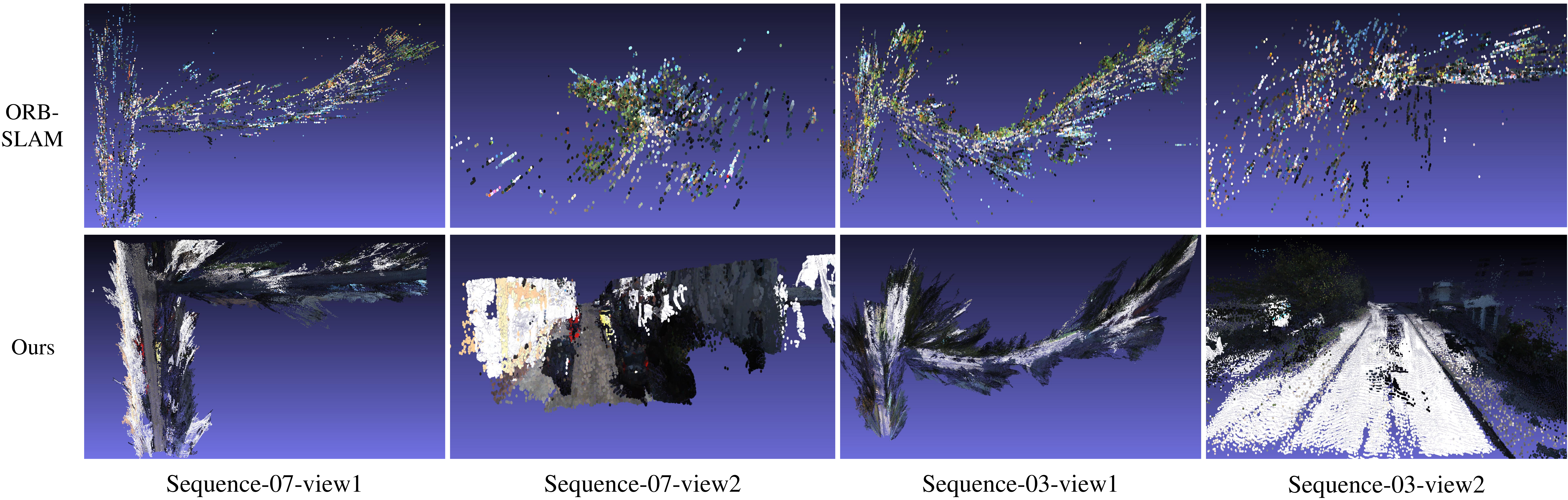}
\end{center}
\caption{Comparison of 3D mapping between ORB-SLAM and ours. As our method can obtain dense depth maps from the monocular depth estimation network, we can reconstruct dense maps with much better quality. It is clear that our mapping can obtain more details of the street. By contrast, ORB-SLAM can only obtain very sparse maps. }
\label{fig:mapping}
\end{figure*}

\begin{table}[]
\centering
\caption{Absolute keyframe trajectory RMSE (in meters) on the RobotCar dataset~\cite{maddern20171}. Trajectories are aligned with 7DoF with the ground truth. `$\times$' denotes that the tracking is lost at some points or a significant portion of the sequence is not processed by the system. Our improvements over the baseline method (ORB-SLAM) are denoted with $\uparrow$.} 
\scalebox{0.9}{
\begin{tabular}{l|cccc}
\toprule[1pt]
\multicolumn{1}{l|}{\multirow{3}{*}{Method}} & \multicolumn{4}{c}{Testing on RobotCar (RMSE)}  \\
\multicolumn{1}{r|}{}   & 2014-05-06   &2014-05-14    &2014-06-25    &2014-05-14  \\ 
                        & -12-54-54   &-13-53-47    &-16-22-15   &-13-59-05 \\  \hline
SVO~\cite{forster2014svo}   & $\times$  & $\times$  & $\times$  & $\times$       \\
DSO~\cite{engel2017direct} & $4.708$  & $\times$   & $\times$  & $2.453$  \\
CNN-SVO~\cite{tateno2017cnn}   & $8.657$  &$6.302$ & $3.703$  & $6.152$  \\
ORB (w/o Loop)~\cite{mur2015orb}  & $10.660$  & $\times$  & $6.558$ & $\times$     \\ \hline
Ours & $6.872^\uparrow$ &$\times$ & $3.563^\uparrow$ & $\times$  \\ \toprule[1pt]
\end{tabular}}
\label{robotcarCmp}
\end{table}

\begin{table}[]
\centering
\caption{Absolute keyframe trajectory error (in centimeters) on the TUM-RGBD dataset~\cite{sturm2012evaluating}. Trajectories are aligned with 7DoF. `$\times$' denotes that the tracking is lost at some points or a significant portion of the sequence is not processed by the system. Our improvements over the baseline method (ORB-SLAM) are noted with $\uparrow$.}
\scalebox{.91}
{
\begin{tabular}{l|cccc}
\toprule[1pt]
\multicolumn{1}{l|}{\multirow{2}{*}{Method}} & \multicolumn{4}{c}{Testing on TUM-RGBD (RMSE)}  \\
\multicolumn{1}{r|}{}   &fr1\_xyz  &fr2\_xyz   & fr3\_sit\_halfsph & fr3\_long\_office  \\  \hline
SVO~\cite{forster2014svo}   & $\times$   &$\times$   & $\times$   & $\times$        \\
DSO~\cite{engel2017direct}  & $11.652$   & $2.377$  & $9.686$  & $24.694$    \\
CNN-SVO~\cite{tateno2017cnn}   & $\times$   & $\times$   & $\times$   & $\times$    \\
ORB (w/o Loop)~\cite{mur2015orb}  & $0.942$   & $0.244$ & $1.657$ & $3.809$   \\ \hline
Ours  & {$0.889^\uparrow$} & {$0.240^\uparrow$}  & {$1.611^\uparrow$} & {$2.496^\uparrow$} \\ \toprule[1pt]
\end{tabular}}
\label{TUMCmp}
\end{table}

\begin{table}[t]
\footnotesize
\renewcommand\arraystretch{1.0}
\begin{center}
\caption{The average processing time (seconds) for each image on KITTI sequence 04. `ORB' represents the `ORB-SLAM'. `DKF' denotes the CNN inference to generate depth for keyframes. `VO' and `VOM' indicates visual odometry and visual odometry with dense mapping respectively. `s/f' represents the average running time to process each frame. 
Note that DKF is performed for keyframes only.}
{
\begin{tabular}{l|c|c|c|c|c}
\toprule[1pt]
Method & CPU & GPU & Frame & Time (s) & Speed (s/f)  \\
\hline
ORB &\checkmark & $\times$ & 271 & 5.24& 0.019  \\
\hline
DKF &\checkmark  & \checkmark & 158 & 65.91 & 0.42  \\
VO &\checkmark  & \checkmark  & 271 & 72.66 & 0.27  \\
VOM &\checkmark  & \checkmark  & 271 & 210.88 & 0.78 \\
\toprule[1pt]
\end{tabular}
}
\label{Table:Speed}
\end{center}
\end{table}

\subsection{Mapping Demonstration}
Previous feature-based VO methods, such as ORB-SLAM~\cite{mur2015orb}, depend on feature matching to find correspondences between two frames to obtain depth. ORB-SLAM has demonstrated that feature-based methods are able to obtain state-of-the-art monocular localization accuracy. However, 
because only a few feature points can be extracted and matched in the feature-based VO, feature-based VO methods are unable to obtain dense maps. 
By contrast, we embed a monocular depth estimation module into a feature-based VO system and use learned monocular depth for mapping. Because the learned depth does not depend on feature matching, our frameworks are able to reconstruct dense maps. The mapping comparison between the baseline and our method is shown in Fig.~\ref{fig:mapping}. As can be seen from Fig.~\ref{fig:mapping}, by using our framework, the feature-based VO can be extended for dense mapping. For the low texture areas, such as the surface of a street, our framework can reconstruct the full shapes.

\subsection{Processing Speed}
We thoroughly evaluate the processing speed of our method on the KITTI dataset. The experiment is performed on Ubuntu 18.04 with an Intel i7-8700 CPU and a GeForce GTX 1060 6GB GPU. In Table~\ref{Table:Speed}, we show that the bottleneck to prevent our method to achieve high processing speed is the inference of monocular depth estimation. For the whole sequence, the total processing time for VO is 72.66 s, while the inference time for the depth estimation network is 65.91 s. More than 90\% of the time is spent on the inference of the depth estimation network.  Compared with ORB-SLAM, although the proposed framework does not have speed advantages, 
we argue that in this work we mainly focus on demonstrating the effectiveness of our framework and do not emphasize the processing speed. In fact, the processing speed can be accelerated by using two approaches. The first approach is simplifying the architecture of our depth estimation network. In this paper,  we use a depth estimation network with a ResNeXt-50 backbone, which is too heavy for platforms with limited computing resources. This can be improved by introducing light network architectures, such as using a ResNet-18~\cite{he2016deep} backbone. 
Another approach is introducing neural network compression~\cite{neill2020overview} or knowledge distillation~\cite{hinton2015distilling} methods to build a more efficient network while maintaining the accuracy.

\begin{table*}[t]
\centering
\caption{Absolute keyframe trajectory RMSE (in meters) on the KITTI dataset~\cite{geiger2012we}. Trajectories are aligned with 7DoF with the ground truth. `$\times$' denotes that the tracking is lost at some points or a significant portion of the sequence is not processed by the system.  $\uparrow$ denotes our improvements to the baseline (DSO).}
\scalebox{1}{
\begin{tabular}{c|ccccccccccc}
\toprule[1pt]
\multicolumn{1}{c|}{\multirow{2}{*}{Method}} & \multicolumn{11}{c}{Testing on KITTI (RMSE)}                                                        \\
\multicolumn{1}{r|}{}                        & $00$       & $01$ & $02$       & $03$     & $04$     & $05$      & $06$      & $07$      & $08$       & $09$      & $10$      \\ \hline
DSO~\cite{engel2017direct}                                          & $113.18$ & $\times$  & $116.81$ & $1.39$ & $0.42$ & $47.46$ & $55.61$ & $16.71$ & $111.08$ & $52.22$ & $11.09$ \\
Ours (DSO)&$87.75^\uparrow$ & $\times$  & $79.97^\uparrow$ & $1.13^\uparrow$ & $0.31^\uparrow$ & $42.81^\uparrow$ & $ 53.66^\uparrow$ & $15.29^\uparrow$ & $89.60^\uparrow$ & $40.29^\uparrow$ & $10.40^\uparrow$ \\
\toprule[1pt]
\end{tabular}}
\label{KITcmp_DSO}
\end{table*}

\begin{table}[t]
\centering
\caption{Absolute keyframe trajectory errors (in centimeters) on the TUM-RGBD dataset~\cite{sturm2012evaluating}. All trajectories are aligned with 7DoF for comparison. 
Our improvements over the baseline method (DSO) are noted with $\uparrow$.}
\scalebox{0.9}{
\begin{tabular}{c|cccc}
\toprule[1pt]
\multicolumn{1}{c|}{\multirow{2}{*}{Method}} & \multicolumn{4}{c}{Testing on TUM-RGBD (RMSE)}  \\
\multicolumn{1}{r|}{}   &fr1\_xyz  &fr2\_xyz   & fr3\_sit\_halfsph & fr3\_long\_office  \\  \hline
DSO~\cite{engel2017direct}  & $11.652$   & $2.377$  & $9.686$  & $24.694$ \\
Ours (DSO) & {$11.149^\uparrow$} & {$2.143^\uparrow$}  & {$6.617^\uparrow$} & {$23.918^\uparrow$} \\ \toprule[1pt]
\end{tabular}}
\label{TUMCmp_DSO}
\end{table}

\subsection{Application in Other VO Frameworks} 
\label{DSO_Exp}
ORB-SLAM is our baseline framework and the proposed method can be used for improving the performance of ORB-SLAM. Compared with other learning-based methods, our framework has the generalization ability to work in diverse scenes. In fact, our method also has a strong extension ability to be applied for other VO frameworks. To illustrate this, we conduct experiments using DSO.
Similar with the application presented in Section~\ref{localization}, we embed our depth estimation module into DSO~\cite{engel2017direct} to investigate the performance. The experiments on the KITTI and TUM-RGBD datasets are shown in Table~\ref{KITcmp_DSO} and Table~\ref{TUMCmp_DSO} respectively. As can be seen from these two tables, the performance of DSO shows obvious improvement on both the KITTI and TUM-RGBD datasets 
with 
introducing our method. For example, our method reduces the RMSE from 116.81 to 79.97 on KITTI sequence 02.
Therefore, the proposed method has the extension ability to be applied in other monocular VO frameworks, not limited to ORB-SLAM.

\section{Conclusion}
In this paper, we have proposed a framework, which combines monocular depth estimation and monocular VO together. 
We have
designed 
a method to train a robust monocular depth estimation network which can perform zero-shot testing in diverse scenes. Our depth estimation network has two working modes: (1) the network takes a single RGB image and outputs a relative depth map; (2) the network takes a single RGB image and a sparse depth map to output high-quality scale consistent dense depth.
The first mode can be used for improving the localization accuracy of VO and the second mode enables our framework to build dense maps. Compared with geometry-based and learning-based methods, our framework is able to obtain better accuracy while maintaining the generalization ability to work in diverse scenes.
Experiments on different indoor and outdoor datasets
demonstrate the effectiveness of our framework.

\section*{Acknowledgments}
We thank the editor and the reviewers for their constructive comments. 

{\small
\bibliographystyle{IEEEtran}
\bibliography{egbib}
}

\end{document}